%% file: sample.tex
\documentclass{opt2020} 
\usepackage{graphicx}
\usepackage{caption}
\usepackage{comment}
\usepackage{multirow}
\SetKwInput{KwInput}{Input}                
\SetKwInput{KwOutput}{Output}              
\SetKwBlock{Loop}{Loop}{end}
\SetKwInput{KwRequire}{Require}
\usepackage{placeins}

\title[A Variant of Gradient Descent Algorithm Based on Gradient Averaging]{A Variant of Gradient Descent Algorithm Based on Gradient Averaging}

\optauthor{
 \Name{Saugata Purkayastha} \Email{sau.pur9@gmail.com}\\
 \addr Assam Don Bosco University, India
 \AND
\Name{Sukannya Purkayastha} \Email{purkayasthasukannya020@gmail.com}\\
\addr Indian Institute of Technology Kharagpur, India
}

\begin{document}

\maketitle

\begin{abstract}%
In this work, we study an optimizer, \texttt{Grad-Avg} to optimize error functions. We establish the convergence of the sequence of iterates of \texttt{Grad-Avg} mathematically to a minimizer (under boundedness assumption). We apply \texttt{Grad-Avg} along with some of the popular optimizers on regression as well as  classification tasks. In regression tasks, it is observed that the behaviour of \texttt{Grad-Avg} is almost identical with Stochastic Gradient Descent (SGD). We present a mathematical justification of this fact. In case of classification tasks, it is observed that the performance of \texttt{Grad-Avg} can be enhanced by suitably scaling the parameters. Experimental results demonstrate that \texttt{Grad-Avg} converges faster than the other state-of-the-art optimizers for the classification task on two benchmark datasets.
\end{abstract}


\section{Introduction} \label{sec:sec1}
Gradient descent (GD) method \cite{lemarechal2012cauchy} is one of the most popular algorithms to optimise an error function. Let $J:\mathbb R^n \rightarrow \mathbb R$  be a $C^1$ function. The classical Gradient Descent method is given by the following algorithm:
\begin{center}
$\theta = \theta-\alpha \nabla J(\theta)$
\end{center}
where $\alpha$ is the constant step size, otherwise known as the learning rate ,$\nabla J(\theta)$ stands for the gradient of the function $J$ and $\theta$ is the set of parameters. 
Although gradient descent algorithm is guaranteed to converge to the global minima (a local minima) for convex functions (non-convex functions), in case of large datasets the convergence can be slow. To overcome this, a variant of GD known as Stochastic Gradient Descent (SGD) \cite{robbins1951stochastic} is applied. SGD, unlike GD avoids redundant computations but has a tendency to overshoot the minima in the process. Both GD and SGD, albeit being efficient in convex optimization, proved to be inefficient in case of non-convex surfaces because of the existence of \emph{saddle points}, as observed in \cite{dauphin2014identifying}. Keeping this in mind two more variants of GD, namely SGD with momentum \cite{ruder2016overview} and Nesterov Accelerated Gradient algorithm \cite{nesterov2013introductory} are commonly used in case of non-convex optimization. One is referred to \cite{ruder2016overview} for a detailed discussion on the above mentioned optimizers. Further, \cite{truong2018backtracking} contains an excellent account of various analytical concepts occuring in the context of machine learning. We define the proposed optimizer, \texttt{Grad-Avg} motivated by the Heun's method \cite{atkinson2008introduction} with the  following iterative scheme:
\begin{equation}
 \theta =\theta -\alpha \frac{1}{2}(\nabla J(\theta)+\nabla J(\theta -\alpha\nabla J(\theta))) \label{eq:1}
\end{equation}
Thus we note that in the present optimizer the update of the parameter is done by considering the average of the gradients calculated at the previous position and the position of the parameter as suggested by the GD algorithm. The intuition lies in the fact that if the gradient becomes zero at a position which is not the minimum (say at a saddle point instead), even then the update of the parameter continues by virtue of the average of the gradients as mentioned above. The concept of gradient averaging is a known concept in this field. For instance \cite{defazio2014saga} introduced a new optimization method using this concept. Also \cite{huang2017snapshot} used the average of model parameters obtained by letting the model to converge at multiple local minimas before making the final prediction . Our work, however uses the averaging of the gradients of the model parameter at every timestep to update the parameter's position.\\
\textbf{Contributions:}
In this work, we develop a new optimizer, \texttt{Grad-Avg} based on gradient descent algorithms. We obtain its convergence for $\alpha \leq \frac{1}{3L}$ where $L$ is the Lipschitz constant (under boundedness assumption). We propose a justification of its similar behaviour with GD for regression tasks. We empirically demonstrate its efficiency over other popular optimizers for classification task.
\section{Assumptions}
In section \ref{sec:sec1}, we have introduced the proposed optimizer. Below we mention the assumptions which are necessary to establish the convergence of the optimizer:
\begin{description}
\item (i) the function  $J:\mathbb R^n \rightarrow \mathbb R$ is $C^1$ i.e. the function $J$ has continuous first order partial derivatives.
\item (ii) the gradient $\nabla J$ of the function  $J:\mathbb R^n \rightarrow \mathbb R$ is Lipschitz continuous i.e. there exists some $L > 0$ (known as the Lipschitz constant) such that $\left\lVert \nabla J(x) - \nabla J(y)\right\rVert < L\left\lVert x - y\right\rVert$ for all $x, y \in \mathbb R^n$.
\end{description}
Further, we make use of Monotone Convergence Theorem which states that every bounded below monotonic decreasing sequence of real numbers converges to the infimum. A detailed account of these concepts can be found in \cite{kreyszig1978introductory} and \cite{shifrin2005multivariable}.
\section{Our Proposed Algorithm}
\begin{algorithm2e}
  \KwData{$(x_i,y_i)$: \text{Input, Output pair}}
  \KwRequire{Information of gradient, $\nabla$ for function, $J$}
  \KwRequire {Learning rate, {$\alpha$}}
  \KwRequire{Initial parameter values, $\theta_{0}$}
  \For {t in range \bf(epochs)}{
$\overline{\theta_{n+1}}= \theta_n - \alpha \cdot \nabla(J(\theta_n);(x_i,y_i))$\;
$\theta_{n+1}=\theta_{n}-\alpha \cdot \frac{\nabla J(\theta_n)+\nabla J(\overline{\theta_{n+1}})}{2}$
}
  \caption{Algorithm for \texttt{Grad-Avg}}
\end{algorithm2e}

\section{Convergence Analysis}
To analyze the convergence of equation \eqref{eq:1}, we first rewrite equation \eqref{eq:1} in the following way:
\begin{equation}
 \theta_{n+1} =\theta_n -\alpha \frac{1}{2}(\nabla J(\theta_n)+\nabla J(\overline{\theta_{n+1}})) \label{eq:2}
\end{equation} 
with 
\begin{equation}
\overline{\theta_{n+1}}= \theta_n -\alpha\nabla J(\theta_n) \label{eq:3}
\end{equation} 
We note that if $\nabla J(\theta_n)=0$, then both $\overline{\theta_{n+1}}$ and $\theta_{n+1}$ reduces to $\theta_n$ i.e. $\theta_n$ is the optimal value. We may thus assume that $\nabla J(\theta_n)\neq 0$ and hence $\left\Vert\nabla J(\theta_n)\right\Vert >0$. We first establish that the sequence $\{J(\theta_n )\}$ where $\{\theta_n \}$ as defined by scheme \eqref{eq:2} is monotonic decreasing in the following result. As such, by the Monotone Convergence Theorem , it follows that $\{J(\theta_n )\}$ converges to the infimum provided it is bounded. 
\begin{theorem}\label{thm:theorem_1}
Let $J:\mathbb R^n \rightarrow \mathbb R$ be a $C^1$ function such that $\nabla J$ is Lipschitz continuous with Lipschitz constant $L$. Then for $\alpha \leq \frac{1}{3L}$the sequence $\{J(\theta_n)\}$ where $\{\theta_n \}$ is defined by scheme \eqref{eq:2} is monotonic decreasing.\\
\end{theorem}
First of all, using the fact that $J:\mathbb R^n \rightarrow \mathbb R$, is Lipschitz continuous, we observe that for $\epsilon_0 >0$, there is some $\delta_0 >0$ such that when $\left\Vert \alpha\nabla J(\theta_n)\right\Vert <\delta_0$, one has
\begin{equation}
\left\Vert \nabla J(\overline{\theta_{n+1}})-\nabla J(\theta_n)\right\Vert <L\delta_0 =\epsilon_0 \label{eq:4}
\end{equation}
Let us now consider $f:\mathbb R \rightarrow \mathbb R$ and define
\begin{equation}
f(t)=J(\theta_n -\alpha t \frac{1}{2}(\nabla J(\theta_n)+\nabla J(\overline{\theta_{n+1}}))) \label{eq:5}
\end{equation}
for $t\in \mathbb R$.
Then by chain rule we get,
\begin{equation}
f^{\prime}(t) =(\nabla J(\theta_n -\alpha t \frac{1}{2}(\nabla J(\theta_n)+\nabla J(\overline{\theta_{n+1}})))) \boldsymbol{\cdot} (-\alpha  \frac{1}{2}(\nabla J(\theta_n)+ \nabla J(\overline{\theta_{n+1}}))) \label{eq:6}
\end{equation}
where $``\boldsymbol{\cdot}"$ stands for the usual dot product.
Now by Fundamental Theorem of Calculus, we get
\begin{eqnarray}
J(\theta_{n+1})-J(\theta_n )&=\int_{0}^{1}(\nabla J(\theta_n -\alpha t \frac{1}{2}(\nabla J(\theta_n)+\nabla J(\overline{\theta_{n+1}}))))\nonumber \\&(-\alpha  \frac{1}{2}(\nabla J(\theta_n)+\nabla J(\overline{\theta_{n+1}})))dt \label{eq:7}
\end{eqnarray}
As by assumption $J$ is $C^1$, hence for $\epsilon = \frac{2\delta}{\alpha t}-(\epsilon_0 +\frac{\delta_0}{\alpha})$ there is a $\delta >0$ such that whenever $\left\Vert \alpha t \frac{1}{2}(\nabla J(\theta_n)+\nabla J(\overline{\theta_{n+1}}))\right\Vert < \delta$, one gets
\begin{equation}
\left\Vert \nabla J(\theta_n -\alpha t \frac{1}{2}(\nabla J(\theta_n)+\nabla J(\overline{\theta_{n+1}})))-\nabla J(\theta_n )\right\Vert <\epsilon \label{eq:8}
\end{equation}
Further, we have the following two inequalities:
\begin{align*}
&\nabla J(\theta_n -\alpha t \frac{1}{2}(\nabla J(\theta_n)+\nabla J(\overline{\theta_{n+1}})))(\nabla J(\theta_n)+\nabla J(\overline{\theta_{n+1}}))\\
&=(\nabla J(\theta_n -\alpha t \frac{1}{2}(\nabla J(\theta_n)+\nabla J(\overline{\theta_{n+1}}))-\nabla J(\theta_n))(\nabla J(\theta_n)+\nabla J(\overline{\theta_{n+1}}))
+ \nabla J(\theta_n)(\nabla J(\theta_n)+\nabla J(\overline{\theta_{n+1}})\\
&\geq\nabla J(\theta_n)(\nabla J(\theta_n)+\nabla J(\overline{\theta_{n+1}}))-\left\Vert \nabla J(\theta_n -\alpha t \frac{1}{2}(\nabla J(\theta_n)+\nabla J(\overline{\theta_{n+1}}))-\nabla J(\theta_n)\right\Vert \left\Vert \nabla J(\theta_n)+\nabla J(\overline{\theta_{n+1}})\right\Vert
\end{align*}
and
\begin{align*}
&\nabla J(\theta_n)(\nabla J(\theta_n)+\nabla J(\overline{\theta_{n+1}}))\\
&=\left\Vert\nabla J(\theta_n)+\nabla J(\overline{\theta_{n+1}})\right\Vert ^2 -\nabla J(\overline{\theta_{n+1}})(\nabla J(\theta_n)+\nabla J(\overline{\theta_{n+1}}))\\
&=\left\Vert\nabla J(\theta_n)+\nabla J(\overline{\theta_{n+1}})\right\Vert ^2 -(\nabla J(\overline{\theta_{n+1}})-\nabla J(\theta_n)+\nabla J(\theta_n))
(\nabla J(\theta_n)+\nabla J(\overline{\theta_{n+1}}))\\
&\geq \left\Vert \nabla J(\theta_n)+\nabla J(\overline{\theta_{n+1}})\right\Vert ^2 -\left\Vert\nabla J(\overline{\theta_{n+1}})-\nabla J(\theta_n)+\nabla J(\theta_n))\right\Vert
\left\Vert(\nabla J(\theta_n)+\nabla J(\overline{\theta_{n+1}}))\right\Vert\\
&\geq \left\Vert \nabla J(\theta_n)+\nabla J(\overline{\theta_{n+1}})\right\Vert ^2 -(\left\Vert\nabla J(\overline{\theta_{n+1}})-\nabla J(\theta_n)\right\Vert+\left\Vert\nabla J(\theta_n))\right\Vert)\left\Vert(\nabla J(\theta_n)+\nabla J(\overline{\theta_{n+1}}))\right\Vert\\
\end{align*}
Now using equation \eqref{eq:8} and the above two inequalities in equation \eqref{eq:7}, one has
\begin{align*}
J(\theta_{n+1})-J(\theta_n ) &\leq -\frac{\alpha}{2}[\frac{4\delta^2}{(\alpha t)^2}-(\epsilon_0 +\frac{\delta_0}{\alpha})\frac{2\delta}{\alpha t}-\epsilon\frac{2\delta}{\alpha t} ] &\leq 0
\end{align*}
Thus $\{J(\theta_n)\}$ is a monotonic decreasing sequence. The proof will be complete once we show that $\epsilon >0$. We note that
\begin{align*}
&\left\Vert \alpha t \frac{1}{2}(\nabla J(\theta_n)+\nabla J(\overline{\theta_{n+1}}))\right\Vert
&\leq \left\Vert \alpha t \frac{1}{2}(\nabla J(\theta_n)\right\Vert + \left\Vert \alpha t \frac{1}{2}\nabla J(\overline{\theta_{n+1}}))\right\Vert 
& <  \frac{1}{2}t\delta_0 + \left\Vert \alpha t \frac{1}{2} \nabla J(\overline{\theta_{n+1}}))\right\Vert 
\end{align*}
i.e. the norm of the sum given by the first expression of the above inequality is bounded by the last expression of the inequality with the minimum being $\frac{t}{2}\alpha_0$ which is attained when $\nabla J(\overline {\theta_{n+1}})=0$. Under the present hypothesis, it is possible only when $\overline{\theta_{n+1}}$ is the minimizer as $\overline{\theta_{n+1}}$ represents the iterates of GD as shown in equation \eqref{eq:7}. Thus, we clearly have $\delta >\frac{t}{2}\alpha_0$ as $\left\Vert \alpha t \frac{1}{2}(\nabla J(\theta_n)+\nabla J(\overline{\theta_{n+1}}))\right\Vert < \delta$. Now simple calculation yields $\epsilon > \frac{1}{\alpha}(\delta_0 -\frac{\delta_0}{2})>\frac{\delta_0}{2\alpha}>0$ since $\alpha <\frac{1}{3L}$ by hypothesis.


\section{Numerical Experiments and Discussion} \label{sec:sec5}
\begin{figure}[!ht]
\centering
\subfigure[MNIST Dataset\label{fig:mnist}]{\includegraphics[scale=0.35]{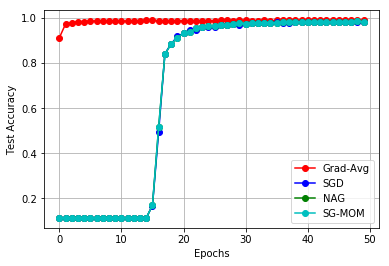}}
\subfigure[MNIST Fashion Dataset\label{fig:mnist_fashion}]{\includegraphics[scale=0.35]{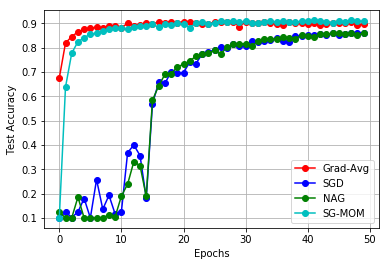}}
\caption{Test set accuracy comparison on MNIST and MNIST Fashion dataset}
\label{fig:classification}
\end{figure}

We perform experiments on two regression datasets, Boston Housing Dataset \cite{harrison1978hedonic} and Condition Based Maintenance of Naval Propulsion Plants Data Set (CBM) \cite{coraddu2016machine}. 
The datasets contain $N$ = 506 and 13 features and $N$ = 11,934 and 15 features respectively where $N$ is the number of datapoints. 
We perform experiments on two different initialization strategies of the neural network weights, the details of which are furnished in Appendix C.
For the classification task, we make use of two of the standard benchmark datasets: the well-known \textbf{MNIST dataset} \cite{lecun2010mnist} and a harder drop-in replacement \textbf{MNIST Fashion dataset}  \cite{xiao2017fashion}. 
Further details about the datasets and the learning models are provided in Appendix A and Appendix B.\\
From Figure \ref{fig:regression} and Table \ref{tab:my_label}, it is clear that the performance of \texttt{Grad-Avg} is almost identical with SGD in case of Regression Tasks. The reason for this lies in the \emph{MSE} loss with which the model is trained in regression tasks. We present a mathematical justification of this phenomena considering a general $C^2$ function represented by a quadratic form. Let $f:\mathbb R^n \rightarrow \mathbb R$ be given by $f(x)=\frac{1}{2}x^ T Qx$. Then $\nabla f=Qx$ and $\nabla^2 f=Q$, the Hessian of $f$. Further, let the $n$ eigenvalues of $Q$ be given by $\lambda_i$ for $i=1, 2, \dots, n$. Then the iterates for Gradient Descent Algorithm for $f$ starting at $x_0$ are given in the following way \cite{lee2016gradient}:
\begin{equation}
x_{t+1}=\sum_{i=1}^{n}(1-\alpha \lambda_i)^{t+1}<e_i ,x_0>e_i \label{eq:9}
\end{equation}
with $e_i$ being the standard basis vectors. The above will certainly converge for $0< \alpha < \frac{2}{L}$ where $L=max|\lambda_i|$ \cite{nesterov2013introductory}. Moreover, Gradient Descent for $f$ converges to a local minimizer or diverges to negative infinity with random initialization depending on whether each critical point of $f$ is a local minimizer or a strict saddle provided $0< \alpha < \frac{1}{L}$ with $L$ being the Lipschitz constant of $\nabla f$, as noted in \cite{lee2016gradient}.
It can be seen by the application of Taylor's theorem that the  iterates of \texttt{Grad-Avg} for $f$ are generated by $x-\alpha \nabla f(x)-O(\alpha^2)$ which for the above bound on $\alpha$, reduces to $x-\alpha \nabla f(x)$ (upon neglecting terms having higher powers of $\alpha$) which is just the SGD scheme. Hence the sequence of iterates of \texttt{Grad-Avg} coincides with equation \eqref{eq:9} above in this case. Now in view of Theorem \ref{thm:theorem_1} and the discussion of the preceding paragraph, it follows that \texttt{Grad-Avg} converges to the local minimizer of $f$ almost identically as SGD. In case of classification tasks, we note that while Figure \ref{fig:classification} indicates that the performance of \texttt{Grad-Avg} is distinct as compared to other optimizers, for MNIST dataset there is yet some scope of improvement. We achieve this passing over to MNIST fashion dataset while keeping the hyper-parameters unchanged as indicated in Figure \ref{fig:classification}. We attribute this improvement to the complexity of the dataset.
\begin{figure}[!ht]
\centering
\subfigure[Boston Housing Dataset\label{fig:boston}]{\includegraphics[scale=0.35]{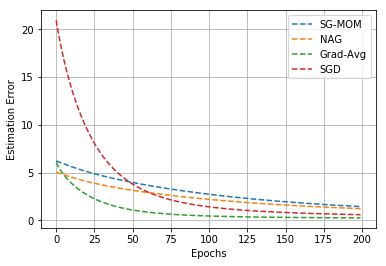}}
\subfigure[CBM dataset\label{fig:cbm}]{\includegraphics[scale=0.35]{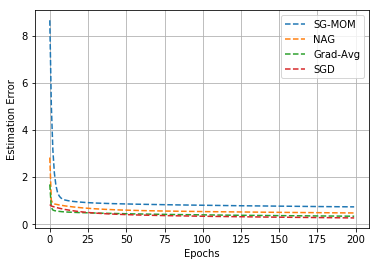}}
\caption{Mean Square Error plots for normal initialization of weights on CBM and Boston Housing Dataset}
\label{fig:regression}

\end{figure}

\begin{table}[ht]
\centering
 \begin{tabular}{c|c|c|c|c}\hline
{Dataset} & {MOM} & {NAG} & \texttt{Grad-Avg} & {SGD} \\ \hline 
      \textbf{Boston Housing} & 14.31 & 2.31 & \textbf{3.35} & 3.36  \\
      \textbf{CBM} & 0.76 & 0.48 & \textbf{0.32} & 0.31 \\ \hline
 \end{tabular}
 \caption{MSE on the test set for both the regression datasets using normal initialization.`MOM' stands for `SGD with Momentum'}
 \label{tab:my_label}
\end{table}

\vspace{0.0 mm}

\section{Conclusion}
 We have  observed that for a regression task, \texttt{Grad-Avg} performs almost identically as SGD while for classification tasks,  \texttt{Grad-Avg} outperforms other optimizers for large datasets applied to complex models. Overall the performance of \texttt{Grad-Avg} is comparable to that of SGD. This together with the fact that SGD surprisingly performs better than most of the standard optimizers irrespective of the task as noted in \cite{luo2019adaptive}, makes \texttt{Grad-Avg} \emph{as good as} SGD. 

\bibliography{sample}
\appendix \label{appendix}
\input{appendix}

\end{document}

%% file: appendix.tex
\section{Datasets} 
We discuss the various datasets used in detail here:
\begin{itemize}
    \item \textbf{Boston Housing Dataset}: The Boston Housing Dataset \cite{robbins1951stochastic} consists of data collected by US Censor Service concerning houses in the areas of Boston Housing mass. The dataset consists of various attributes such as per capita-income,  proportion of non-retail business acres per town, etc. The price of a particular house given these attributes is to be predicted.
    \item \textbf{CBM Dataset}: Condition Based Maintenance of Naval Propulsion Plants Data Set (CBM) \cite{coraddu2016machine} consists of data from experiments carried out by a numerical simulator of a naval vessel (Frigate) characterized by a Gas Turbine (GT) propulsion plant. The GT propulsion plant is characterised by various components such as, Lever position, ship speed, Gas Turbine Shaft Torque, etc. Based on these attributes, the Gas Turbine Compressor Decay co-efficient is to be calculated.
\end{itemize}
The datasets were split into $80\%-20\%$ randomly for training and testing. For the classification task, we make use of two of the standard benchmark datasets: the well-known \textbf{MNIST dataset} \cite{lecun2010mnist} and a harder drop-in replacement \textbf{MNIST fashion dataset}  \cite{xiao2017fashion}. Both of these datasets consist of 28*28 grayscale images. Each image needs to be classified into one of the $10$ classes. For the experiments, we have used the pre-defined train-test split that consists of $60,000$ images for training and $10,000$ images for testing.

\section{Hyper-parameters and Learning Models}
\begin{itemize}
    \item \textbf{Regression Task}: We use a single layer Neural Network as our learning model for the experiments. The model is parameterized by a set of weights, $w$ and bias, $b$. For all the experiments, we use the same set of hyper-parameters. The learning rate is set to be $5e-5$, momentum is set at $0.9$ for SG-MOM and NAG, batch size is set to be $50$ and the models are trained for $200$ epochs.
    \item \textbf{Classification Task}: We utilize a Convolutional Neural Network (CNN) as our learning model. The CNN consists of a convolutional layer with kernel size of $3$ pixels having a $20$ channel output. The output is followed by a max-pooling layer of $2$ pixels. The output is then passed through a convolution layer of $5$ pixels having a $50$ channel output, followed by a max-pooling layer of the same configuration as before. The output is then flattened and connected to a layer of $128$ neurons which is connected to a $10$-neuron output layer.We train the model for $50$ epochs with a learning rate of $0.01$ for all the optimizers. The momentum value is set at $0.9$. The batch size is set to be $128$.
\end{itemize}
\section{Error Plots}
\begin{figure}
\centering
\subfigure[Boston Housing Dataset\label{fig:uni_boston}]{\includegraphics[scale=0.35]{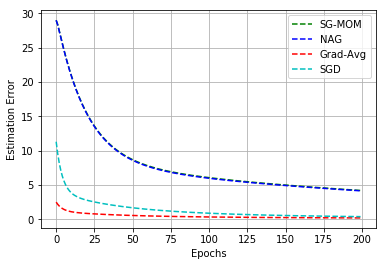}}
\subfigure[CBM Dataset\label{fig:uni_cbm}]{\includegraphics[scale=0.35]{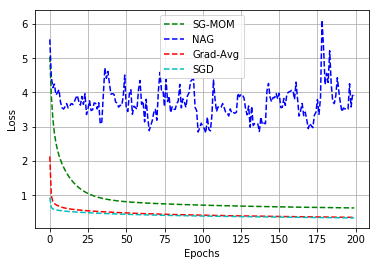}}
\caption{Mean Square Error Plots on Boston Housing and CBM dataset}
\label{fig:Uniform}
\end{figure}

We show the error plots obtained for the Uniform Initialization of Weights for the Regression task on Boston Housing and CBM dataset in Figure \ref{fig:Uniform}.
\begin{figure}
\centering
\subfigure[MNIST dataset\label{fig:MNIST_loss}]{\includegraphics[scale=0.35]{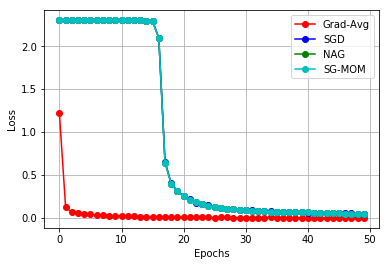}}
\subfigure[MNIST Fashion Dataset\label{fig:MNIST_fashion_loss}]{\includegraphics[scale=0.35]{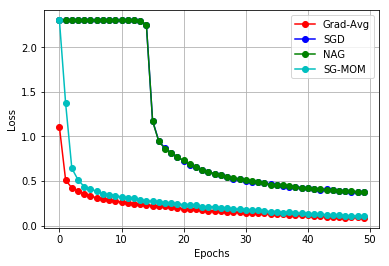}}
\caption{Error plots on MNIST and MNIST Fashion Dataset}
\label{fig:Loss}
\end{figure}
\begin{table}
\centering
 \begin{tabular}{c|c|c|c|c}\hline
{Dataset}
    & {SG-MOM} & {NAG} & \texttt{Grad-Avg} & {SGD} \\ \hline 
      \textbf{Boston Housing} & 38.60 & 8.70 & 6.64 & \textbf{6.61}  \\
      \textbf{CBM} & 2.42 & 1.22 & \textbf{0.51} & \textbf{0.51}  \\ \hline
 \end{tabular}
 \caption{MSE on the test set for both the regression datasets for uniform initialization.`SG-MOM' stands for `SGD with Momentum'}
 \label{tab:my_label_2}
\end{table}
\vspace{0.0 mm}

Thus it is clear that \texttt{Grad-Avg} behaves identically with SGD in case of Uniform Initialization too on the same pair of datasets as considered under Normal Initialization. The justification which we have provided in section \ref{sec:sec5} for Normal Initialization remains valid in this case also as the underlying error function is MSE. Further, owing to the same reason it may be noted that the convergence of \texttt{Grad-Avg} to a local minimizer is almost guaranteed (in the sense of \cite{lee2016gradient}) irrespective of the choice of initialization.\\
Finally from the error plots, on the classification task as shown in Figure \ref{fig:Loss} on the same pair of datasets as considered in section \ref{sec:sec5},  it is evident that \texttt{Grad-Avg} converges faster to the minimizer as compared to the other optimizers for the training set which further validates our claim of faster convergence of \texttt{Grad-Avg}. 